%% file: main.tex
\documentclass[10pt,twocolumn,letterpaper]{article}

\usepackage{3dv}
\usepackage{times}
\usepackage{epsfig}
\usepackage{graphicx}
\usepackage{amsmath}
\usepackage{amssymb}
\usepackage{multirow}
\usepackage{makecell}
\usepackage{xcolor}
\usepackage{subcaption}

\usepackage{comment}

\usepackage[pagebackref=true,breaklinks=true,colorlinks,bookmarks=false]{hyperref}

\include{math_commands}

\newcommand{\misscite}{\textcolor{red}{[C]}}

\threedvfinalcopy 

\def\threedvPaperID{267} 

\ifthreedvfinal\fi
\begin{document}

\title{GAN2X: Non-Lambertian Inverse Rendering of Image GANs \vspace{-0.1cm}}

\author{
	Xingang Pan\textsuperscript{\normalfont 1}
	\and
	Ayush Tewari\textsuperscript{\normalfont 1,2}
	\and
	Lingjie Liu\textsuperscript{\normalfont 1}
	\and
	Christian Theobalt\textsuperscript{\normalfont 1}
	\\
	\vspace{-0.5cm}
	\and
	\textsuperscript{\normalfont 1}{\normalfont Max Planck Institute for Informatics} \
	\qquad\quad\textsuperscript{\normalfont 2}{\normalfont MIT} \\
	\qquad\tt\small \{xpan,lliu,theobalt\}@mpi-inf.mpg.de  \quad ayusht@mit.edu
	\\
	{\normalsize Project page: \href{https://vcai.mpi-inf.mpg.de/projects/GAN2X/}{https://vcai.mpi-inf.mpg.de/projects/GAN2X/}}
}

\twocolumn[{
	\renewcommand\twocolumn[1][]{#1}
	\maketitle
	\begin{center}
		\centering
		\vspace{-0.95cm}
		\includegraphics[width=\linewidth]{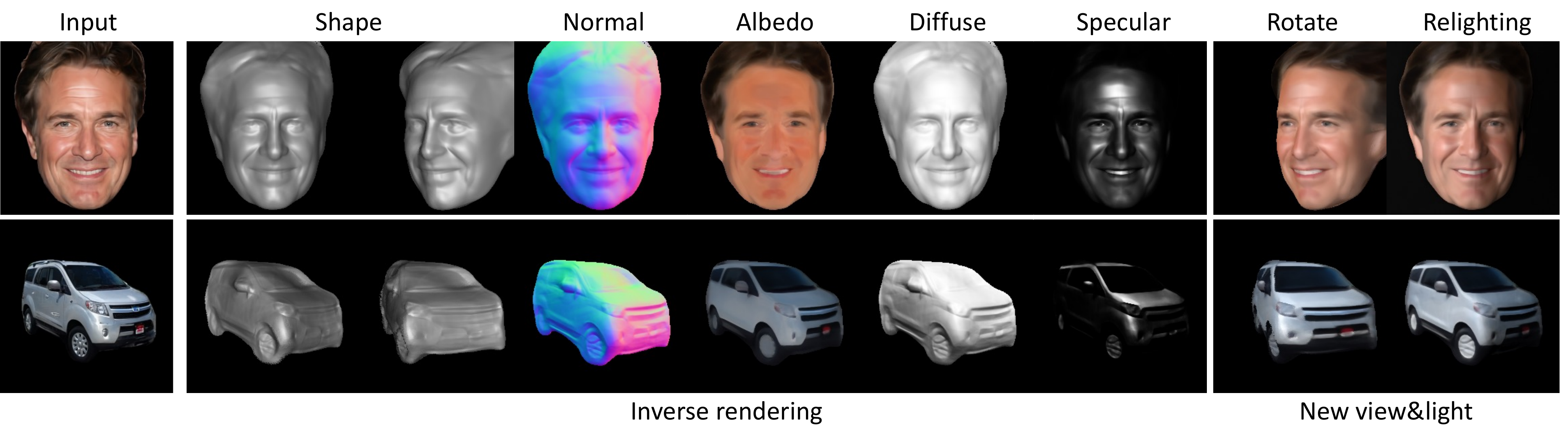}
		\vspace{-0.75cm}
		\captionof{figure}{We study non-Lambertian inverse rendering of image GANs. For an image generated by a GAN pretrained on an unpaired image collection, we leverage its corresponding pseudo paired images to recover the underlying object shape, albedo, specular properties, and lighting. This allows to re-render the object under a novel viewpoint or lighting condition.  
		}
		\label{fig:teaser}
	\end{center}
}]

\maketitle

\input{sections/abstract}
\input{sections/introduction}
\input{sections/relatedwork}
\input{sections/method}
\input{sections/experiments}

\input{sections/conclusion}

\section*{Acknowledgements}
\vspace{-0.1cm}
Christian Theobalt was supported by ERC Consolidator Grant 4DReply (770784). Lingjie Liu was supported by Lise Meitner Postdoctoral Fellowship.
We thank Abhimitra Meka for providing the Total Relighting results.

{\small
\bibliographystyle{ieee_fullname}
\bibliography{bib}
}

\input{sections/appendix.tex}

\end{document}

%% file: math_commands.tex
\usepackage{amsmath,amsfonts,bm}

\def\eqref#1{equation~\ref{#1}}

\def\1{\bm{1}}

\newcommand{\test}{\mathcal{D_{\mathrm{test}}}}

\def\vtheta{{\bm{\theta}}}
\def\va{{\bm{a}}}

\def\vd{{\bm{d}}}

\def\vh{{\bm{h}}}

\def\vl{{\bm{l}}}

\def\vn{{\bm{n}}}
\def\vo{{\bm{o}}}

\def\vr{{\bm{r}}}

\def\vv{{\bm{v}}}
\def\vw{{\bm{w}}}
\def\vx{{\bm{x}}}

\def\vz{{\bm{z}}}

\def\mA{{\bm{A}}}

\def\mC{{\bm{C}}}

\def\mL{{\bm{L}}}

\DeclareMathAlphabet{\mathsfit}{\encodingdefault}{\sfdefault}{m}{sl}
\SetMathAlphabet{\mathsfit}{bold}{\encodingdefault}{\sfdefault}{bx}{n}

\DeclareMathOperator*{\argmin}{arg\,min}

%% file: sections/abstract.tex

\begin{abstract}

2D images are observations of the 3D physical world depicted with the geometry, material, and illumination components.
Recovering these underlying intrinsic components from 2D images, also known as inverse rendering, 
 usually requires a supervised setting with paired images collected from multiple viewpoints and lighting conditions, which is resource-demanding. 
In this work, we present GAN2X, a new method for unsupervised inverse rendering that only uses unpaired images for training.
Unlike previous Shape-from-GAN approaches that mainly focus on 3D shapes, we take the first attempt to also recover non-Lambertian material properties by exploiting the pseudo paired data generated by a GAN.
To achieve precise inverse rendering, we devise a specularity-aware neural surface representation that continuously models the geometry and material properties.
A shading-based refinement technique is adopted to further distill information in the target image and recover more fine details.
Experiments demonstrate that GAN2X can accurately decompose 2D images to 3D shape, albedo, and specular properties for different object categories, and achieves state-of-the-art performance for unsupervised single-view 3D face reconstruction.
We also show its applications in downstream tasks including real image editing and lifting 2D GANs to decomposed 3D GANs.

\end{abstract}

%% file: sections/introduction.tex

\section{Introduction}








Natural images are formed as a function of the objects' underlying physical attributes, such as geometry, material, and illumination.
Estimating these object intrinsics from images is one of the core problems in computer vision, and is often referred to as inverse rendering, \ie, the inverse process of rendering in computer graphics.
It has wide applications in AR/VR and visual effects, such as relighting, material editing, and object pose editing.

A classic approach for inverse rendering is photometric stereo~\cite{alldrin2008photometric,goldman2009shape}, which requires multi-view and multi-lighting images of a scene to be captured with a light stage.
These paired images provide sufficient information for estimating object intrinsics.
However, the need of a sophisticated light stage setup makes it difficult to apply to diverse object categories, especially in-the-wild objects like cars.
To get rid of this constraint, another line of work aims at  performing inverse rendering in an unsupervised or weakly-supervised manner~\cite{kanazawa2018learning,kanazawa2018end,goel2020shape,wu2020unsupervised}, where only unpaired 2D image collections or weak annotations are given.
Despite great progress, these approaches often miss fine-grained geometry details and cannot recover specular properties due to the inherent ambiguity. 


\begin{figure}[t!]
	\centering
	\includegraphics[width=\linewidth]{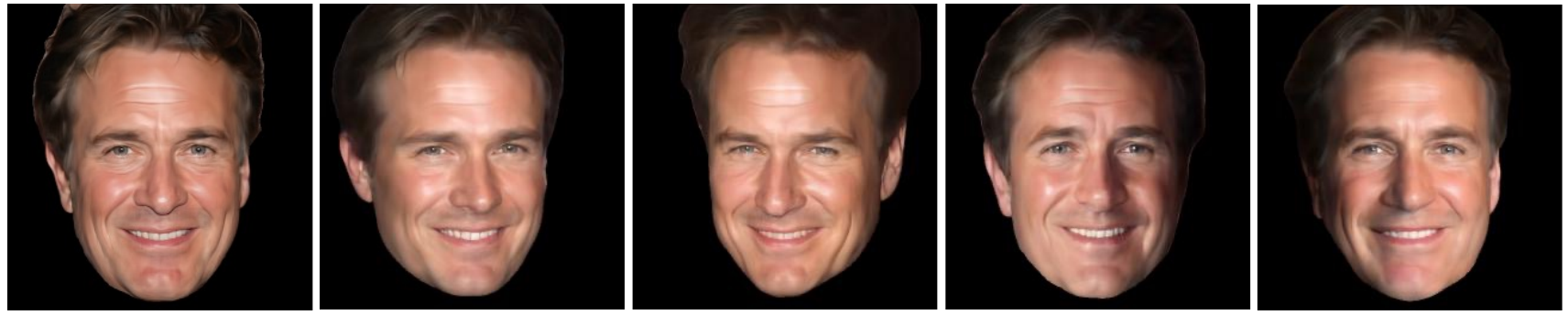}
	\vspace{-0.6cm}
	\caption{Images generated by a GAN~\cite{karras2020analyzing} with different viewpoints and lighting conditions. The various specular highlight effects reveal the non-Lambertian material properties of the face. }
	\vspace{-0.4cm}
	\label{fig:GAN_specular}
\end{figure}

While the lack of paired data is the main challenge of unsupervised inverse rendering, 
a continuous image distribution modelled by a GAN~\cite{goodfellow2014generative,karras2020analyzing} provides a possibility to create pseudo light-stage data.
GANs implicitly mimic the image formation process by learning from unpaired image collections.
It is shown in the recent Shape-from-GAN works \cite{pan2020gan2shape,StyleGAN3D} that the pseudo paired data generated by GAN can be used to reconstruct 3D shapes. 
However, these methods do not account for specular material properties and their mesh-based representation limits the precision of reconstructed 3D shapes.
We believe that exploiting GANs for inverse rendering is still not fully explored.
For instance, Fig.~\ref{fig:GAN_specular} provides GAN-generated images with variations on not only viewpoints but also the specular highlights, showing evidence that GANs implicitly capture the spatially-varying specular properties of the face.


Motivated by the above observation, in this work, we are interested to go one step further, exploring the problem of non-Lambertian inverse rendering of image GANs with finer details. 
In particular, we present a method that, for each image generated by a pretrained GAN, learns to predict its underlying 3D shape, material properties including albedo and specular components (\ie, specular intensity and shininess), and lighting condition.
Our method partially uses \cite{pan2020gan2shape} as a tool, but we go beyond it to further explore if GANs can also be used to disentangle material properties.
While the training is based on GAN-generated images, our method can also be applied to real images. 

Concretely, to model high-quality 3D shape and non-Lambertian appearance, we devise an implicit neural surface representation to model the 3D surface, albedo, and specular properties. 
Rendering images from these intrinsics are conducted based on differentiable volume rendering~\cite{kajiya1984ray} and Phong shading~\cite{phong1975illumination}.
Inspired by \cite{pan2020gan2shape}, we adopt an exploration-and-exploitation algorithm to generate pseudo multi-view and multi-lighting images from a pretrained GAN.
The neural surface representation can then be optimized with a reconstruction loss on the pseudo paired data along with regularization losses that facilitate training.
However, the multi-view images generated by 2D GANs are not strictly 3D consistent, leading to imprecise results with noticeable mismatch to the target image. 
To address this issue, we adopt a shading-based refinement process that adapts our model to fully exploit information from the target image and effectively resolves the mismatch and recovers more fine details. 

With the above designs, our approach, namely GAN2X, can 
reconstruct high-quality 3D shape, albedo, and specular properties for different object categories,
as shown in Fig.~\ref{fig:teaser}.
The results show that image GANs do implicitly capture non-Lambertian object intrinsics, and thus provide a new way towards unsupervised inverse rendering.
With such capability, GAN2X is also a powerful approach for unsupervised 3D shape learning from unpaired image collections. 
On the task of albedo and surface normal estimation and single-view 3D face reconstruction, GAN2X significantly outperforms existing unsupervised baselines.
We also show promising results in downstream applications including image relighting and lifting 2D GANs to 3D with decomposed intrinsics.

Our contributions are summarized as follows: 
(1) We propose GAN2X that achieves high-quality non-Lambertian inverse rendering using GANs 
pretrained on unpaired images only. 
Our work reveals the large potential of GANs for learning spatially-varying non-Lambertian material properties.
(2) GAN2X naturally serves as a strong unsupervised 3D shape  learning approach. We show the state-of-the-art performance on unsupervised single-view 3D reconstruction of faces.
(3) GAN2X enables a wide range of downstream applications including lifting 2D GANs to 3D with decomposed intrinsics and photorealistic visual effects like relighting and material editing.



%% file: sections/relatedwork.tex

\section{Related Work}

\subsection{Generative Adversarial Networks}

We have witnessed the great progress of Generative Adversarial Networks (GANs) in image synthesis starting from the first GAN model~\cite{goodfellow2014generative}. 
In particular, the StyleGAN family~\cite{karras2019style,karras2020analyzing,Karras2021} has successfully produced photo-realistic and high-resolution imaginary.
However, the image synthesis process of GANs is usually treated as a black box, which lacks physical interpretability (\eg, 3D shape, material properties).
There has been some works that extend GANs to enable 3D control 
~\cite{tewari2020stylerig,deng2020disentangled,zhou2020rotate,tewari2020pie,FreeStyleGAN2021,sarkar2021style,sarkar2021humangan}, 
but such control is driven by the guidance of an external 3D morphable model~\cite{blanz1999morphable} or a 3D mesh input. 
Different from these methods, our method does not require additional 3D geometry as input and can explicitly recover the physical attributes including 3D shape and material properties.

Another line of work is 3D-aware GANs, which adopt 3D representations (\eg, voxel grids, and neural fields) or their integration with 2D generative models to enable 3D controllable image synthesis~\cite{schwarz2020graf,chan2021pi,niemeyer2021campari,gu2021stylenerf,Chan2021,zhou2021CIPS3D,xu2021generative,pan2021shadegan,xu2022volumegan,tewari2022d3d}.
However, most of these methods do not explicitly model the reflectance and the shading process, thus resulting in suboptimal 3D geometry and the lack of control on lighting. 
While~\cite{pan2021shadegan} models illumination explicitly, it makes the assumption of Lambertian reflectance, which does not account for specular highlight.
Besides, the learned 3D shapes in \cite{pan2021shadegan} still lack fine-grained details due to limited resolution for training.
In contrast, we make the first attempt to recover non-Lambertian material properties from pretrained GANs.
Unlike previous works that adopt memory-consuming 3D representations to learn a 3D GAN, our work shows that the 3D geometry and material are also readily achievable from off-the-shelf 2D GANs that are more efficient to train.




\subsection{Supervised Inverse Rendering}
%
Inverse rendering has been well studied given paired images of an object from multiple viewpoints and light conditions.
%
A typical way is to estimate object intrinsics by fitting photometric or geometric models to reconstruct the paired images.
Conventional photometric stereo methods~\cite{alldrin2008photometric,goldman2009shape,luan2021unified} perform this based on meshes while recently it is extended to implicit neural representations~\cite{boss2021nerd,srinivasan2021nerv,zhang2021nerfactor,boss2021neuralpil,zhang2021physg}. 
These methods are object-specific, \ie, they do not generalize to unseen object instances.
Some learning-based methods are also generalizable by learning from multi views, enabling test-time reconstruction with sparse views or a single view~\cite{yu2019inverserendernet,bi2020deep,boss2020two,yu2020self,barron2014shape,kulkarni2015deep}.
%
%
%
%
%
However, collecting multi-view and multi-lighting images is resource-demanding and difficult to apply to in-the-wild objects like cars.
Some methods use synthetic data for training~\cite{li2018learning,meka2018lime,sengupta2018sfsnet,li2020inverse,sang2020single}, but training on such data requires us to solve the domain-gap problem for generalization to real in-the-wild images.
%
To get rid of these limitations, 
in this work, we focus on the unsupervised setting where only unpaired image collections are available for training.

\subsection{Unsupervised Inverse Rendering}

In contrast to the supervised approaches, there has been a surge of interest to develop inverse rendering methods trained only on unpaired image collections.
These methods predict the 3D geometry and appearance reconstruction from the input image and use image-space losses computed between the input and the rendered reconstruction. 
Since this is an ill-posed problem, early approaches focused on human faces and bodies using 3D priors~\cite{tewari2017mofa,kanazawa2018end}, with some methods also learning components of the prior from videos~\cite{tewari2019fml,tewari2021learning}.
Several methods learn to reconstruct the object shapes of general categories~\cite{goel2020shape,kanazawa2018learning,ye2021shelf} using weak supervision like template shapes, masks, or hand-crafted priors (\eg, object symmetry and smoothness).
%
There are a few attempts for learning shape as well as material and illumination from unpaired image collections. 
Wu et al. \cite{wu2020unsupervised} uses object symmetry and assumes Lambertian materials. 
The follow-up work~\cite{wu2021derender} does not limit the object to be Lambertian, but requires objects to have rotational symmetry. 
Similar to us, \cite{pan2020gan2shape} and \cite{StyleGAN3D} also exploit GANs to reconstruct 3D shapes.
GAN2Shape~\cite{pan2020gan2shape} assumes Lambertian reflectance and does not recover high-quality albedo, while \cite{StyleGAN3D} does not disentangle albedo and illumination.
Concurrent to us, \cite{wimbauer2022rendering} relies on coarse 3D shape initialization to estimate the object intrinsics, and assumes globally shared specular intensity and shininess.
%
In contrast, our approach does not rely on coarse shape, and recovers spatially-varying non-Lambertian material properties.
In this work, we show that GANs implicitly capture object shape and material properties, and thus provide a new way for unsupervised inverse rendering with precise recovery of geometry and material properties.


%% file: sections/method.tex

\section{Method}

\begin{figure*}[t!]
	\centering
	\includegraphics[width=\linewidth]{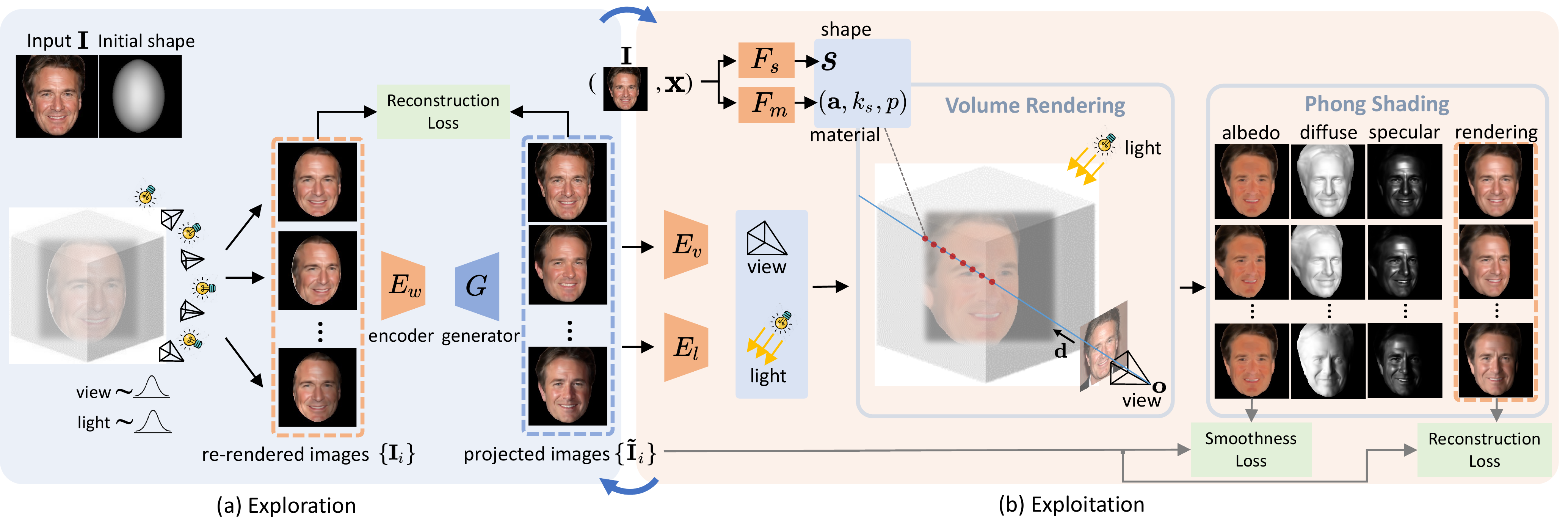}
	\vspace{-0.5cm}
	\caption{\textbf{Method Overview.} 
	(a) In the exploration step, we use a convex shape prior to guide the GAN generator to produce projected images with various viewpoint and lighting conditions. 
	(b) In the exploitation step, we leverage the projected images as pseudo paired data to optimize the underlying intrinsic components. The intrinsic components including shape, albedo, and specular properties are represented via implicit neural fields. Images can be rendered from this representation via volume rendering and Phong shading, which are naturally amenable to gradient-based optimization. }
	\vspace{-0.2cm}
	\label{fig:pipeline}
\end{figure*}

Our approach aims to recover the intrinsic components (3D shape, albedo, specular properties) at high quality for any image generated by an image GAN.
In the following, we first provide some preliminaries on GANs.
We then describe how we represent the intrinsic components and render images from them.
Finally, we introduce our inverse rendering algorithm that distills the knowledge of the GAN to recover the intrinsic components.


\noindent\textbf{Preliminaries on Generative Adversarial Networks (GANs).}
A GAN learns the data distribution via a min-max game between a generator \textit{G} and a discriminator \textit{D}~\cite{goodfellow2014generative}.
After training from an image dataset, the generator can map a latent code $\vz$ to an image.
In this work, our study is based on StyleGAN2~\cite{karras2020analyzing}, a classic GAN consisting of two parts.  
First, the latent code $\vz \in \mathcal{Z}$ is mapped to an intermediate latent code $\vw \in \mathcal{W}$ via a mapping network $F_w : \mathcal{Z} \mapsto \mathcal{W}$.
Then $\vw$ is used to produce the output image with a synthesis network.
We refer to the synthesis network as $G$ in the following sections.

\subsection{Shape and Material Model}
\label{sec:representation}

\textbf{Scene representation.} 
In order to model object shape and material with high fidelity, we adopt implicit neural fields to represent these factors.
Specifically, for any image $\mathbf{I} \in \mathbb{R}^{3 \times H \times W}$ generated by the GAN, its 3D shape is represented by an MLP $F_s: (\mathbf{I},\vx) \mapsto s$, where $\vx \in \mathbb{R}^3$ is the coordinate of any point and $s \in \mathbb{R}$ is its signed distance $s$ to the object surface.
Thus, the surface $\mathcal{S}$ of the object is $\mathcal{S} = \{\vx \in \mathbb{R}^3 | F_s(\mathbf{I}, \vx) = 0 \}$.
The material properties are represented by another MLP $F_m: (\mathbf{I}, \vx) \mapsto (\va, k_s, p)$, where $\va \in \mathbb{R}^3$ is the diffuse albedo, $k_s \in [0, 1]$ is the specular intensity, and $p \in [p_{min}, p_{max}]$ is the shininess.
Unlike NeRF~\cite{mildenhall2020nerf} that models view-dependent color, in our formulation the material property is view-independent while view-dependent effects are explicitly modeled with shading as would be introduced later.
%
%
The conditioning of $F_s$ and $F_m$ on $\mathbf{I}$ is implemented by concatenating the embedding of $\mathbf{I}$ with $\vx$ as the input.
The embedding can be obtained from 1) an additional encoder $E_I$ and 2) the latent code $\vw$ corresponding to $\mathbf{I}$ from the GAN.
Their differences will be discussed latter.

Apart from $F_s$ and $F_m$, we also have a viewpoint encoder $E_v$ that takes $\mathbf{I}$ as input and predicts the camera view $\vv \in \mathbb{R}^{6}$ and a lighting encoder $E_l$ that predicts the lighting condition $\mL = (\vl, k_a, k_d)$.
Here $\vl \in \mathbb{S}^2$ is the light direction, $k_a \in [0,1]$ is the ambient coefficient, and $k_d \in [0,1]$ is the diffuse coefficient.
This scene representation allows to render an image $\mathbf{\hat{I}}$ from any viewpoint or lighting condition via a rendering process $\Phi$ as $\mathbf{\hat{I}} =\Phi(F_s(\mathbf{I}), F_m(\mathbf{I}), \vv, \mL)$.
Next, we introduce $\Phi$ in detail.

\noindent\textbf{Rendering.} 
Solving the inverse rendering problem requires our model to render images from the above intrinsics in a differentiable and easy-to-optimize manner.
We use volume rendering~\cite{kajiya1984ray} and Phong shading~\cite{phong1975illumination} to achieve this.
To render the color $\mC$ of a camera ray $\vr(t) = \vo + t \vd$ with near and far bounds $t_n$ and $t_f$, we first render the albedo $\mA$, specular intensity $K_s$, shininess $P$, and surface normal $\vn \in \mathbb{R}^3$ via:
\begin{align}
\mA(\vr), K_s(\vr), P(\vr) &= \int_{t_n}^{t_f} w(t) F_m(\vr(t)) dt \label{eq:render} \\
\bm{n}(\vr) &= \bm{\hat{n}}(\vr) / \| \bm{\hat{n}}(\vr) \|_2, \nonumber \\
\text{where} ~ \bm{\hat{n}}(\vr) = &- \int_{t_n}^{t_f} w(t) \nabla_{\vr(t)}F_s(\vr(t)) dt,
\label{eq:normal}
\end{align}
Here $w(t)$ is the weight function for volume rendering, for which we use the normalized S-density following NeuS~\cite{wang2021neus}.
%
%
We then perform Phong shading to get the final color $\mC$:
\begin{align}
\mC = \tau \Big(k_a \mA + k_d \left(\text{max}({0, \vn \cdot \vl}) \mA + K_s \text{max}({0, \vn \cdot \vh})^P\right) \Big),
\label{eq:shading}
\end{align}
where $\vh$ is the bisector of the angle between viewing direction $-\vd$ and $\vl$, and $\tau(\vx) = \vx ^ {1/\gamma}, \gamma=2.2$ is a tone-mappping function to ensure a more even brightness distribution.
A limitation of this vanilla Phong shading is that it produces similar darkness at surface areas opposite to the light source.
We observe that the surface with smaller $\vn \cdot \vl$ often tends to be darker due to complex reflection in the environment.
Thus, we optionally replace $\text{max}({0, \vn \cdot \vl})$ with $\text{max}({0, \vn \cdot \vl}) + k_n \text{min}({0, \vn \cdot \vl}), k_n \in [0,1]$, where the new negative term is to compensate for the darkness variation at opposite-light areas.
We show that this revision leads to more realistic relighting in experiments.

Note that $w(t)$ in Eq.~\ref{eq:render} and Eq.~\ref{eq:normal} has a learnable inverse standard deviation parameter $\sigma$ ($s$ in~\cite{wang2021neus}) that controls the concentration of density on the surface.
It is directly optimized during training in~\cite{wang2021neus}.
However, in our case, this would produce a sub-optimal solution as the 3D inconsistency of GAN generated images would impede the convergence of $1/\sigma$.
Thus, we manually increase $\sigma$ from $\sigma_{min}$ to $\sigma_{max}$ with an exponential schedule as $\sigma_i = \sigma_{max} + \text{exp}(-i \beta) (\sigma_{min} - \sigma_{max})$, where $i$ is the number of training iterations and $\beta$ controls the decay.

\subsection{Inverse Rendering}
\label{sec:inverse_render}

%
%
We perform inverse rendering based on the shape and material model introduced above.
In order to generate a number of approximated paired images of various viewpoint and lighting conditions using the GAN, we adopt an exploration and exploitation algorithm following \cite{pan2020gan2shape}.
Different from \cite{pan2020gan2shape}, in this work the inverse rendering is based on the new non-Lambertian neural representation and rendering equation introduced before.
Besides, we also devise 
a chromaticity-based smoothness loss to regularize the material and a viewpoint and lighting loss to stabilize training.
A shading-based refinement technique will also be introduced in the next subsection.

\noindent\textbf{Exploration.} 
We first initialize $F_s$ to produce an ellipsoid as a convex shape prior following \cite{pan2020gan2shape}.
The camera viewpoint $\vv$ and lighting $\mL$ are initialized to be a canonical setting $\vv_0$ and $\mL_0$.
We then optimize the material network $F_m$ using the reconstruction loss $\mathcal{L}(\mathbf{I}, \Phi(F_s(\mathbf{I}), F_m(\mathbf{I}), \vv_0, \mL_0))$, where $\mathcal{L}$ is L1 loss.
%
As shown in Fig.~\ref{fig:pipeline} (a), with this initial guess of the scene, we can re-render a number of new images $\{\mathbf{I}_i | i = 1,2,...,m\}$ from different viewpoints $\{\vv_i\}$ and lighting conditions $\{\mL_i\}$ using $\Phi$, where $\{\vv_i\}$ and $\{\mL_i\}$ are randomly sampled from their prior distributions.
These re-rendered images $\{\mathbf{I}_i\}$ roughly reveal the change of viewpoint and lighting, thus can serve as a guidance for the exploration in the GAN image manifold.
Specifically, we reconstruct $\{\mathbf{I}_i\}$ using the GAN generator $G$ by training an encoder $E_w$ that predicts the latent codes. 
%
After training, this would produce the GAN-reconstructed images $\{\tilde{\mathbf{I}}_i = G(E_w(\mathbf{I}_i))\}$, namely projected images, which have viewpoints and lighting conditions that resemble the re-rendered images $\{\mathbf{I}_i\}$.
%
%

\noindent\textbf{Exploitation.} 
The projected images $\{\tilde{\mathbf{I}}_i\}$ obtained in the exploration process could be viewed as pseudo paired images for $\mathbf{I}$.
Thus, we exploit their information to recover the intrinsic components.
As shown in Fig.~\ref{fig:pipeline} (b), the viewpoint $\tilde{\vv}_i$ and lighting $\tilde{\vl}_i$ for each image $\tilde{\mathbf{I}}_i$ are predicted by the viewpoint encoder $E_v$ and lighting encoder $E_l$ respectively.
They are then used to render images with the shared shape field and material field.
The scene related networks $F_s$, $F_m$, $E_v$, $E_l$ can be jointly optimized using the reconstruction loss:
\begin{align}
\mathcal{L}_{recon} = \frac{1}{m} \sum_{i=0}^{m} \mathcal{L}^{'}\Big(\tilde{\mathbf{I}}_{i}, \Phi\big(F_s(\mathbf{I}), F_m(\mathbf{I}), E_v(\tilde{\mathbf{I}}_{i}), E_l(\tilde{\mathbf{I}}_{i})\big)\Big) \label{eq:reconstruct}
\end{align}
where $\mathcal{L}^{'}$ is a combination of L1 loss and perceptual loss. 

In addition, we also regularize the material using a chromaticity-based smoothness loss.
For the material maps ($\mA, K_s, P$) rendered from ($F_s(\mathbf{I}), F_m(\mathbf{I}), E_v(\tilde{\mathbf{I}}_{i}), E_l(\tilde{\mathbf{I}}_{i})$), we denote their concatenation as $\mathbf{M}_i$.
The chromaticity-based smoothness loss is defined as:
\begin{align}
    \mathcal{L}_{reg} = \frac{1}{m} \sum_{i=0}^{m} \big( \|\psi(\delta_x \mathbf{\Gamma}_i) \delta_x \mathbf{M}_i\|_1 + \|\psi(\delta_y \mathbf{\Gamma}_i) \delta_y \mathbf{M}_i\|_1 \big) \label{eq:smooth}
\end{align}
where $\mathbf{\Gamma}_i$ is calculated from $\tilde{\mathbf{I}}_{i}$ by taking its $a^*b^*$ channels in the CIELAB color space,
$\delta_x$ and $\delta_y$ are used to signify the computation of image gradients, and $\psi(\vx) = \text{exp} (-\|\vx\|^2/\gamma), \gamma=10$ is a non-linear function.
This regularization motivates pixels with close chromaticity to have similar materials.

As the projected images share similar viewpoints and lighting conditions as the re-rendered images, we can use $\{\vv_i\}$ and $\{\mL_i\}$ to guide the learning of $E_v$ and $E_l$ at the beginning of training:
\begin{align}
    \mathcal{L}_{vl} = \frac{1}{m} \sum_{i=0}^{m} \big(\|\{\vv_i\} - E_v(\tilde{\mathbf{I}}_{i})\|_1 + \|\{\vl_i\} - E_l(\tilde{\mathbf{I}}_{i})\|_1 \big)  \label{eq:viewlight}
\end{align}
In summary, the final training objective is:
\begin{align}
    \vtheta_s, \vtheta_m, \vtheta_v, \vtheta_l = \argmin_{\vtheta_s, \vtheta_m, \vtheta_v, \vtheta_l} \mathcal{L}_{recon} + \lambda_{reg} \mathcal{L}_{reg} + \lambda_{vl} \mathcal{L}_{vl} \label{eq:loss}
\end{align}
where $\vtheta_s, \vtheta_m, \vtheta_v, \vtheta_l$ are the parameters of networks $F_s, F_m, E_v, E_l$ respectively, and $\mathcal{L}_{vl}$ is only used in the first 1k training iterations.

With this training objective, the object shape, material, and lighting will be more correctly inferred.
We then use these updated intrinsic components as the initialization and repeat the exploration and exploitation steps for a few times.
This allows to further distill the knowledge of the GAN and refine the results.

\noindent\textbf{Joint training.} 
We have introduced how our method can be applied to an individual instance $\mathbf{I}$ generated by the GAN.
In this case, the dependence of $F_s$ and $F_m$ on $\mathbf{I}$ is actually not necessary.
Having this dependence allows us to further extend our method for joint training on multiple instances as done in \cite{pan2020gan2shape}, which improves generalization.
For example, if the conditioning of $F_s$ and $F_m$ on $\mathbf{I}$ is achieved by training an additional image encoder $E_I$, then we can apply our model to real images after joint training. 
And if the conditioning is based on the latent code $\vw$ in the $\mathcal{W}$ space of the GAN, then the StyleGAN mapping network $F_w$ together with our $F_s$ and $F_m$ forms a 3D GAN with decomposed intrinsic components.
We show the applications of these variants in experiments.

\begin{figure*}[t!]
	\centering
	\includegraphics[width=17cm]{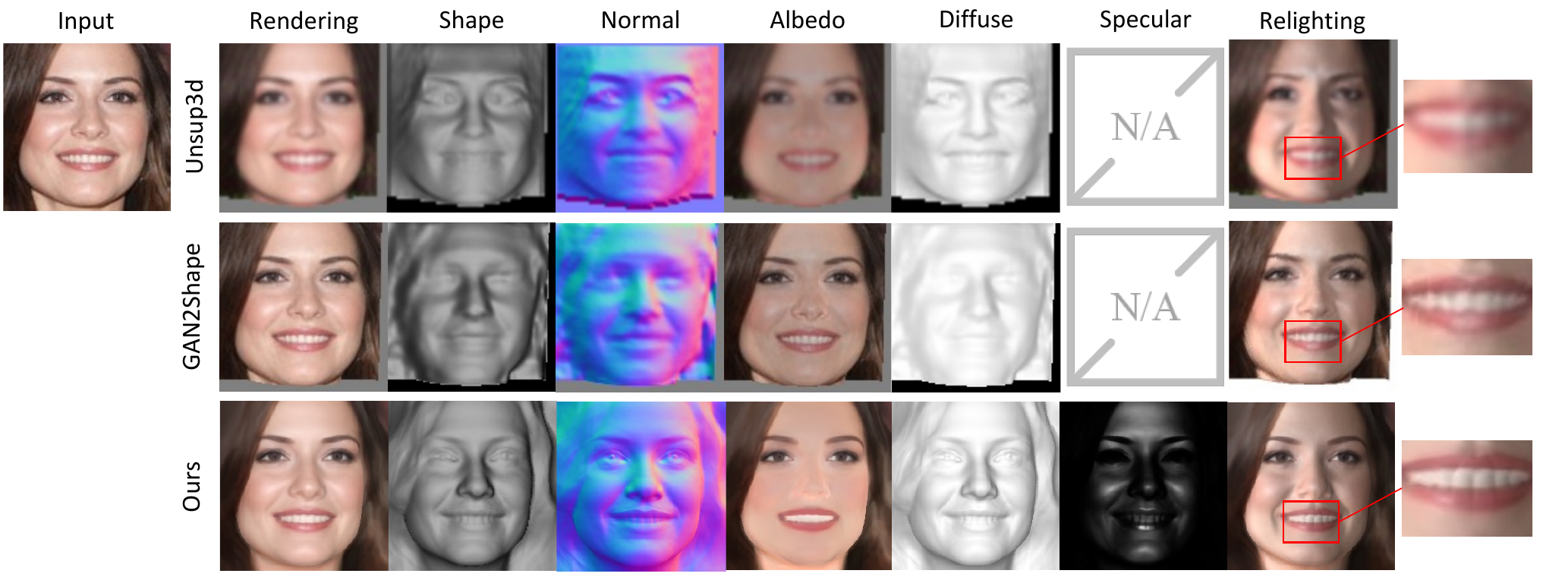}
	\vspace{-0.2cm}
	\caption{\textbf{Qualitative comparison.} Our approach achieves more accurate inverse rendering than baselines (zoom in to see details). Our relighting result successfully models the shift of specular highlight on the lip while baselines cannot.}
	\vspace{-0.1cm}
	\label{fig:compare}
\end{figure*}

\subsection{Shading-based Refinement}

While the above approach already produces pleasing results, the reconstructed shape and appearance could possibly bear noticable mismatch with the target image. 
This is because the pseudo paired data generated by a 2D GAN is not strictly 3D-consistent.
To this end, we draw inspiration from shape-from-shading literature~\cite{zollhofer2015shading} and adopt a shading-based refinement (SBR) step that fully exploits the information in the target image $\mathbf{I}$.
Let $\hat{\mathbf{I}}$, $\hat{\mathbf{D}}$, and $\hat{\mathbf{M}}$ denote the image, depth, and material map rendered from ($F_s(\mathbf{I}), F_m(\mathbf{I}), E_v(\mathbf{I}), E_l(\mathbf{I})$).
The shading-based refinement objective is to reconstruct the target image with several constraints:
\begin{align}
    \vtheta_s, \vtheta_m, \vtheta_v, \vtheta_l &= \argmin_{\vtheta_s, \vtheta_m, \vtheta_v, \vtheta_l} \mathcal{L}^{'}(\mathbf{I}, \hat{\mathbf{I}}) + \lambda_d \| \mathbf{D}, \hat{\mathbf{D}} \|_1 \nonumber \\
    &+ \lambda_m \| \mathbf{M}, \hat{\mathbf{M}} \|_1 + \lambda_{reg} \mathcal{L}_{reg} \label{eq:refinement}
\end{align}
where $\mathbf{D}$ and $\mathbf{M}$ are the initial values of $\hat{\mathbf{D}}$ and $\hat{\mathbf{M}}$ and are used to prevent the shape and material from deviating too much.
Here $\mathcal{L}_{reg}$ is defined only for $\hat{\mathbf{M}}$.
This optimization goal resolves the mismatch and recover more details from the target image while preserving a valid 3D shape and material, as we will show in experiments.

%% file: sections/experiments.tex

\section{Experiments}

We conduct extensive experiments to evaluate our approach GAN2X on unsupervised inverse rendering across different object categories including human faces, cat faces, and cars.
We use datasets of unpaired image collections including CelebA~\cite{liu2015deep}, CelebAHQ~\cite{karras2018progressive}, AFHQ cat~\cite{choi2020starganv2}, and LSUN car~\cite{yu2015lsun}.
The GAN models we used are StyleGAN2~\cite{karras2020analyzing} pretrained on these datasets.
We use off-the-shelf scene parsing models~\cite{yu2018bisenet,zhao2017pyramid} to remove the background.
We recommend readers to refer to the supplementary material for more
implementation details, qualitative results, and videos.

\subsection{Inverse Rendering Results} \label{sec:results}

\noindent\textbf{Qualitative evaluation.}
Fig.~\ref{fig:compare} shows the qualitative comparison between our approach and two unsupervised inverse rendering baselines Unsup3d~\cite{wu2020unsupervised} and GAN2Shape~\cite{pan2020gan2shape} on the CelebA dataset.
We visualize the rendering, 3D shape, surface normal, albedo, diffuse light map, and specular light map.
Our approach reconstructs much more precise 3D shapes than the baselines, \eg, we successfully recover the double-fold eyelids, teeth, and most wrinkles while the baselines miss fine details and produce blurry shapes.
Besides, the baselines cannot decompose the specular light map. Consequently, the specular highlights are entangled into albedo maps.
In contrast, our approach successfully disentangles the specular highlights from the diffuse albedo, \eg, the highlights on the lips are correctly captured in the specular light map.
This, in turn, results in a more accurate albedo map and produces more realistic relighting effects.

\begin{figure*}[t!]
	\centering
	\includegraphics[width=\linewidth]{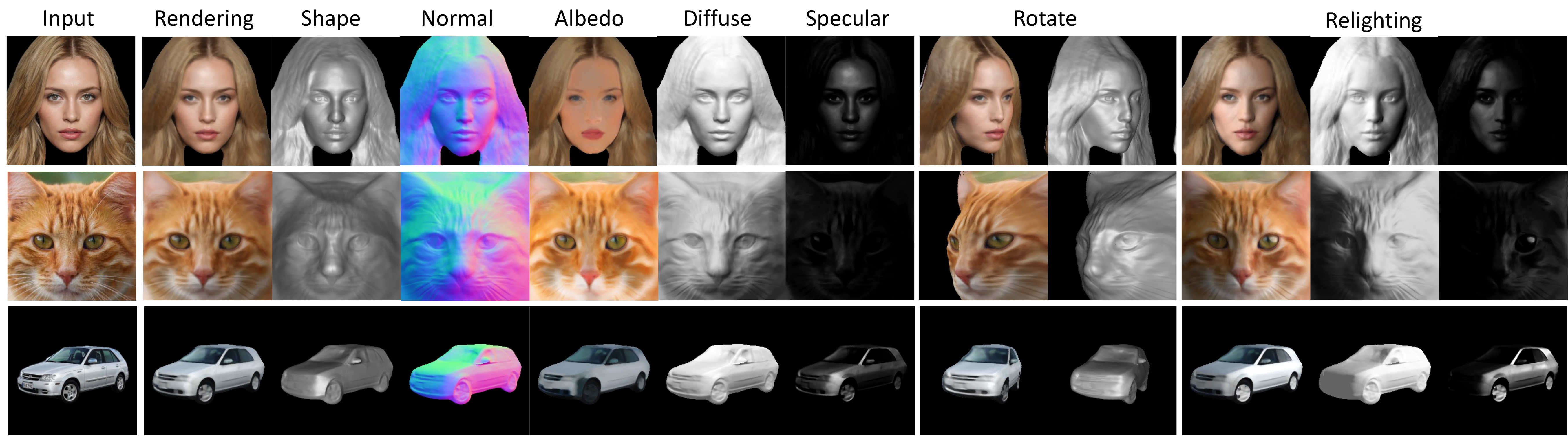}
	\vspace{-0.5cm}
	\caption{\textbf{Qualitative results.} We show inverse rendering and image editing results on the CelebaHQ, AFHQ Cat, and LSUN Car datasets.}
	\vspace{-0.2cm}
	\label{fig:qualitative}
\end{figure*}

\begin{figure*}[t!]
	\centering
	\begin{minipage}{0.4\linewidth}
	\centering
\includegraphics[width=6.5cm]{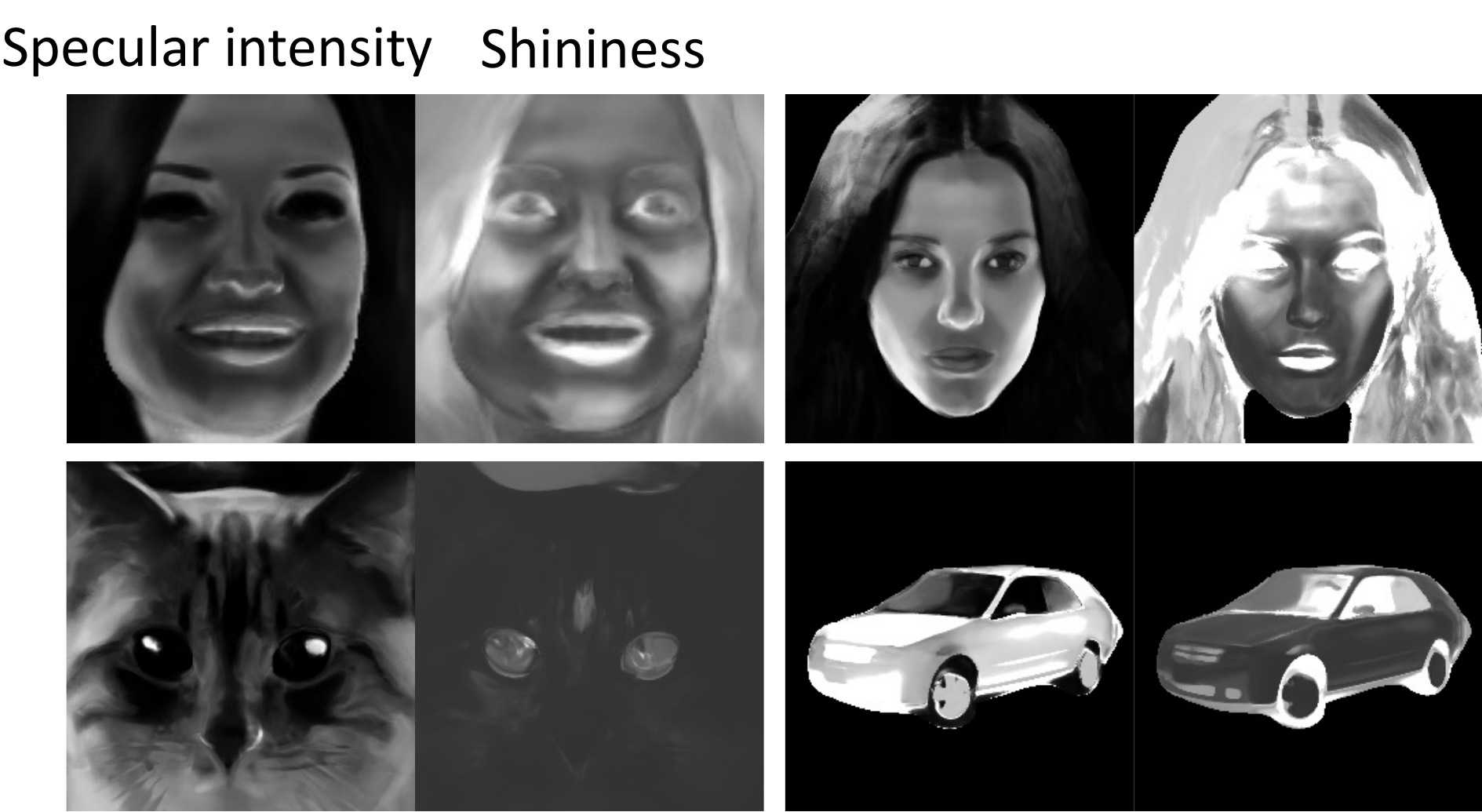}
\vspace{-0.1cm}
\caption{\textbf{Visualization of specular components} (\eg, specular intensity and shininess) for objects in Fig.~\ref{fig:compare} and Fig.~\ref{fig:qualitative}.}
\vspace{-0.1cm}
\label{fig:specular}
	\end{minipage}
	\hfill
	\begin{minipage}{0.57\linewidth}
		\centering
	\includegraphics[width=10cm]{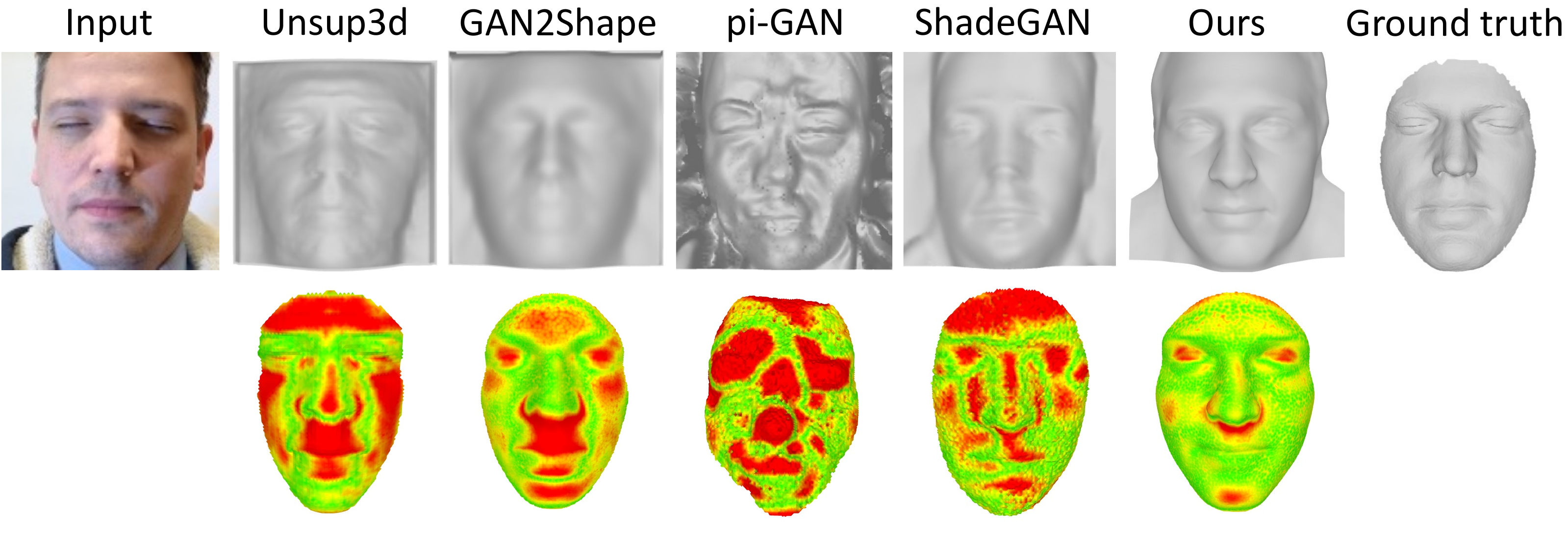}
\vspace{-0.5cm}
\caption{\textbf{Qualitative comparison} of single-view 3D reconstruction on H3DS dataset.}
\vspace{-0.1cm}
\label{fig:h3ds}
	\end{minipage}
	\vspace{-0.1cm}
\end{figure*}

Our results on more datasets are shown in Fig.~\ref{fig:qualitative}.
GAN2X performs high-quality inverse rendering for human head, cat face, and cars, where the 3D shape, albedo, diffuse light map, and specular light map are accurately recovered.
Besides, these reconstructed intrinsic components allow photorealistic re-rendering of the object under a novel viewpoint or lighting condition, showing large potential for image editing. 
We also visualize the learned specular components of different objects in Fig.~\ref{fig:specular}.
It is observed that our model successfully learns some meaningful object-varying and spatially-varying specular properties. 
For instance, the human lips have higher shininess than the human faces; the eyes of the cat have higher shininess than its fur; the car has higher specular intensity than other objects.
These results verify our assumption that GANs implicitly capture object material properties.
We note that learning spatially-varying specular intensity and shininess is very challenging and our method does not resolve all the ambiguities.
%
For example, the tires of the car have high shininess, which is counter-intuitive.
This is because its specular intensity is already very low and thus the shininess cannot get supervision.
Addressing these ambiguities is left as a future work.

\noindent\textbf{Quantitative evaluation.}
%
For quantitative evaluation, we first perform single-view 3D reconstruction on the H3DS dataset~\cite{ramon2021h3d} to evaluate the 3D shape.
The H3DS test set has ground truth 3D face scans for several identities.
We take a single-view image as input and report the Chamfer Distance between our reconstructed shape and the ground truth.
Apart from Unsup3d and GAN2Shape, we also compare with pi-GAN~\cite{chan2021pi} and ShadeGAN~\cite{pan2021shadegan} that can perform unsupervised 3D reconstruction via GAN inversion.
All methods are trained on the CelebA dataset.
Applying our model to these real images is achieved with the jointly trained image encoder $E_I$ and $F_s$ as discussed in Sec.~\ref{sec:inverse_render}.
As shown in Fig.~\ref{fig:h3ds} and Tab.~\ref{tab:h3ds}, our method reconstructs more accurate and more natural-looking 3D shapes than other baselines.
Thus, our approach is also well suited for unsupervised 3D shape learning.

\begin{table}[t!]
\centering
\resizebox{8.5cm}{0.7cm}{
\begin{tabular}{c|cccccc}
\Xhline{2\arrayrulewidth}
\hspace{-0.2cm}Method\hspace{-0.1cm} & \hspace{-0.1cm}Unsup3d\hspace{-0.2cm} & GAN2Shape\hspace{-0.2cm} & pi-GAN\hspace{-0.2cm} & ShadeGAN\hspace{-0.3cm} & \begin{tabular}[c]{@{}c@{}}Ours\\ w/o SBR\end{tabular} \hspace{-0.2cm} & Ours\hspace{-0.2cm} \\ \hline\hline
CD $\downarrow$     & 3.60    & 2.62      & 3.29   & 2.49     & 2.21 & \textbf{2.08} \\ \Xhline{2\arrayrulewidth}
\end{tabular}
}
\vspace{-0.2cm}
\caption{\textbf{Single-view 3D reconstruction} on the H3DS dataset. We report chamfer distance (CD) between the predicted mesh and the ground truth mesh.}
\vspace{-0.2cm}
\label{tab:h3ds}
\end{table}

Evaluating albedo for our GAN-based inverse rendering setting is challenging as there lacks a suitable dataset with ground truth albedo.
To this end, we leverage the state-of-the-art supervised approach Total Relighting~\cite{pandey2021total} to produce pseudo ground truth, and report results on it as a reference. 
Total Relighting is trained on high-quality light-stage data, thus producing stable and reliable albedo and surface normal for human faces.
We test on 500 images generated by the GAN trained on CelebA and report the scale-invariant error (SIE) for albedo and mean-angle deviation (MAD) for surface normal following~\cite{wimbauer2022rendering}.
As shown in Tab.~\ref{tab:albedo}, our approach significantly outperforms baselines, showing better capability to recover face albedo and shape.

\begin{table}[t!]
\centering
\resizebox{7.0cm}{0.72cm}{
\begin{tabular}{c|cccc}
\Xhline{2\arrayrulewidth}
          & Unsup3d~ & GAN2Shape  & Ours \\ \hline\hline
SIE ($\times 10^{-2}$) $\downarrow$   & 3.21   & 3.05 & \textbf{2.16} \\
MAD $\downarrow$  & 18.66   & 21.75 & \textbf{12.67} \\ \Xhline{2\arrayrulewidth}
\end{tabular}
}
\vspace{-0.2cm}
\caption{\textbf{Quantitative comparison} of albedo and surface normal on CelebA. We report scale-invariant error (SIE) for albedo and mean-angle deviation (MAD) for surface normal. }
\vspace{-0.3cm}
\label{tab:albedo}
\end{table}

\begin{figure}[t!]
	\centering
	\includegraphics[width=\linewidth]{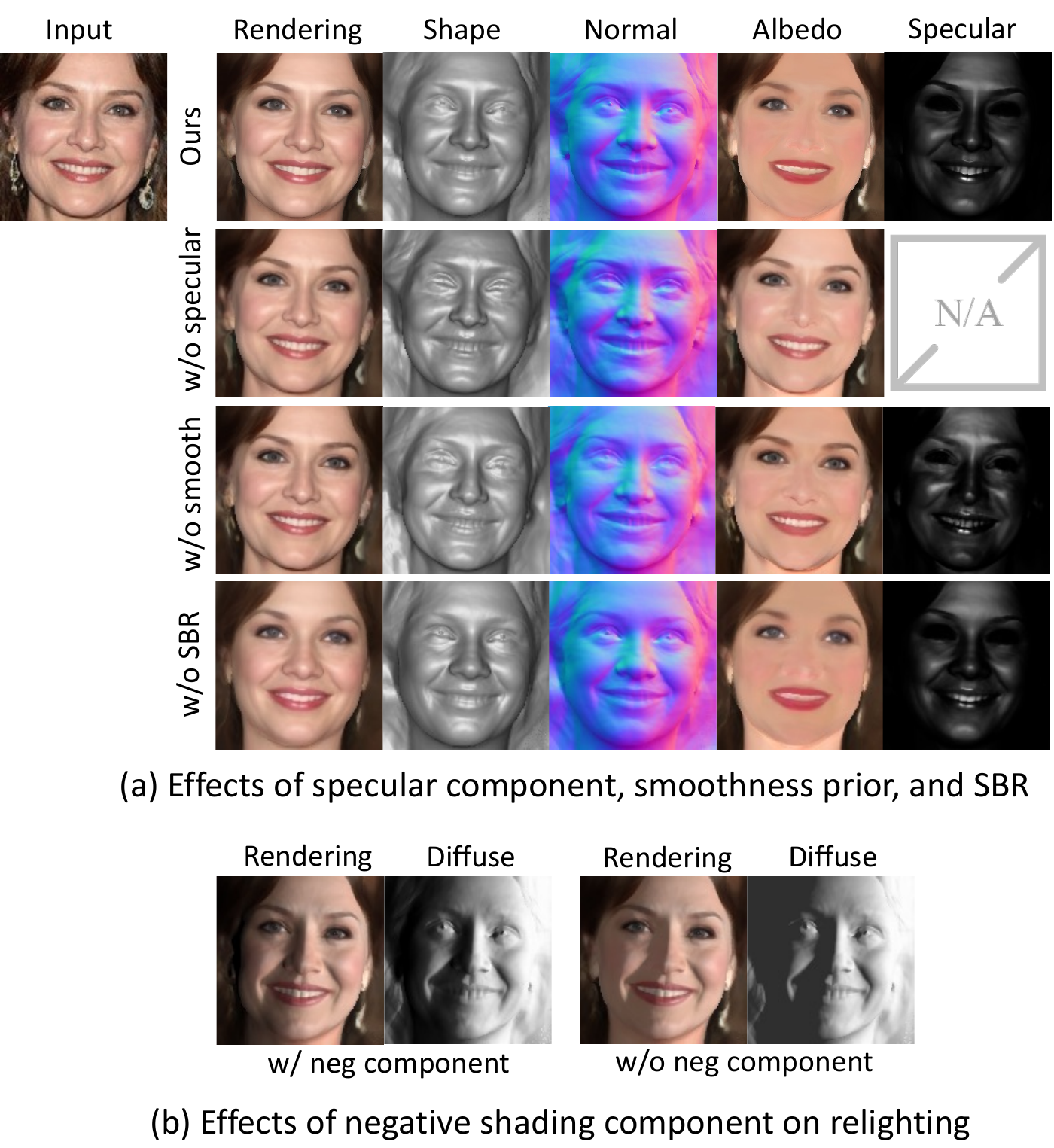}
	\vspace{-0.5cm}
	\caption{\textbf{Ablation study} of (a) specular component, smoothness prior, shading-based refinement (SBR) and (b) negative shading component.}
	\label{fig:ablation}
\end{figure}

\noindent\textbf{Ablation study.}
%
We further study the effects of several components in our pipeline, including the specular term in Eq.~\ref{eq:shading}, the chromaticity-based smoothness loss in Eq.~\ref{eq:smooth}, the shading-based refinement (SBR), and the negative shading term introduced in Sec.~\ref{sec:representation}.
The results are shown in Tab.~\ref{tab:ablation} and Fig.~\ref{fig:ablation}.
It can be observed that including the specular term helps  disentangle specular highlight with albedo. The reduced ambiguity in turn facilitates the learning of surface normal.
Besides, the smoothness prior also leads to better disentanglement between albedo and shading, producing more natural albedo map.
The example of Fig.~\ref{fig:ablation} (a) is a challenging case where the rendering of our approach without SBR exhibits noticable mismatch with the input image. The SBR can effectively reduce the mismatch and recover more details in the target image.
Quantitative results also confirm that SBR further improves the shape and albedo in our approach.
%
Finally, the negative shading term has little effect on the quantitative results, but would improve qualitative relighting effects as it can compensate for darkness variation at surfaces opposite to the light source. 
As shown in Fig.~\ref{fig:ablation} (b), the rendering without this negative shading term tends to produce constant darkness at the opposite-light areas while including this term reveals some geometry details and is more realistic.

\begin{table}[t!]
	\centering
	\resizebox{8.5cm}{0.7cm}{
		\begin{tabular}{c|ccccc}
			\Xhline{2\arrayrulewidth}
			& \hspace{-0.2cm}Ours\hspace{-0.3cm} & w/o specular\hspace{-0.2cm} & w/o smooth\hspace{-0.2cm} & w/o SBR\hspace{-0.2cm} & w/o neg\hspace{-0.2cm} \\ \hline\hline
			\hspace{-0.2cm}SIE ($\times 10^{-2}$) $\downarrow$ \hspace{-0.25cm}  & 2.16   & 2.25 & 2.28 & 2.19 & 2.15 \\
			MAD $\downarrow$  & 12.67   & 12.88 & 12.63 & 12.74 & 12.66 \\ \Xhline{2\arrayrulewidth}
		\end{tabular}
	}
		\vspace{-0.2cm}
		\caption{\textbf{Ablation study} of several design choices. The values have the same meaning as in Tab.~\ref{tab:albedo}. }
	\vspace{-0.1cm}
	\label{tab:ablation}
\end{table}



\subsection{Other Applications}

\begin{figure}[t!]
	\centering
	\includegraphics[width=\linewidth]{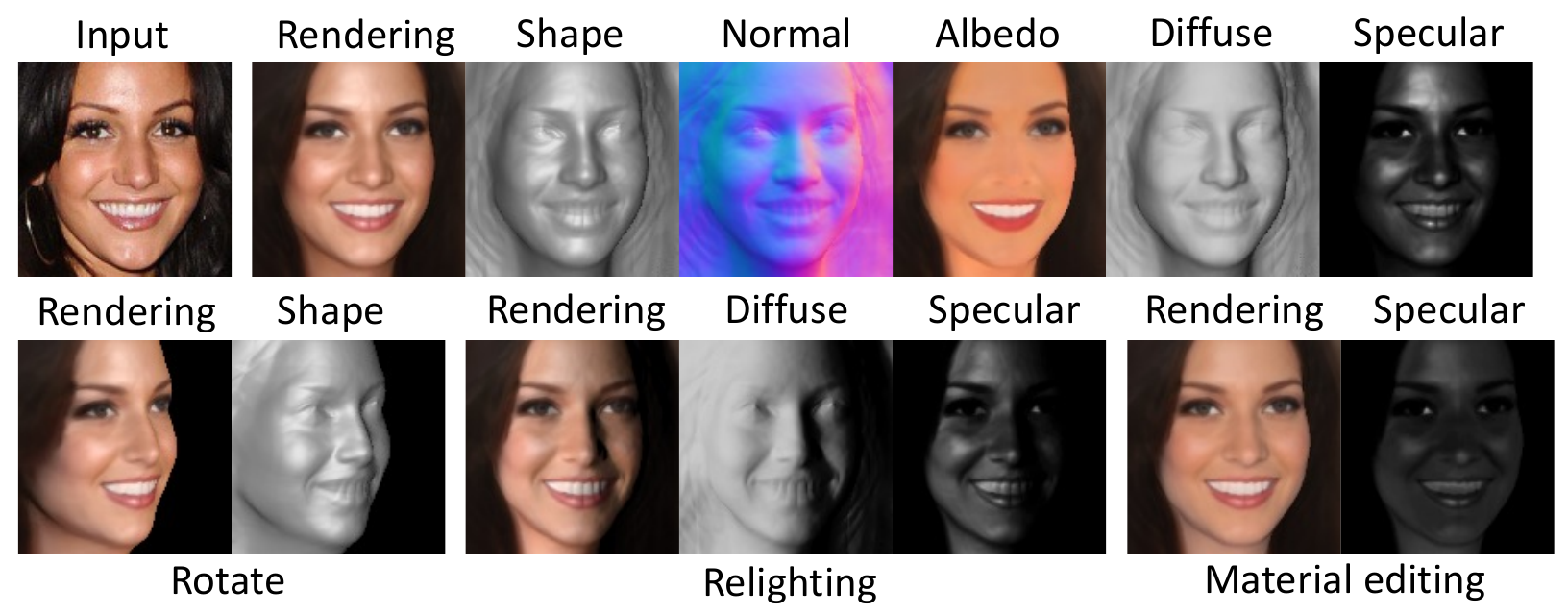}
	\vspace{-0.75cm}
	\caption{Application on real image inverse rendering and editing.}
	\vspace{-0.35cm}
	\label{fig:editing}
\end{figure}

\noindent\textbf{Real image editing.}
We have shown that GAN2X can be used for single-view 3D reconstruction of real images, here we also demonstrate its applicability for other editing effects for real images.
Fig.~\ref{fig:editing} provides an example of a real image, where our method successfully achieves satisfactory inverse rendering and image editing results.
Here we also show material editing effects enabled by our method.
By reducing the specular intensity and shininess, we can weaken the specular highlight on the face.

\begin{figure}[t!]
	\centering
	\includegraphics[width=\linewidth]{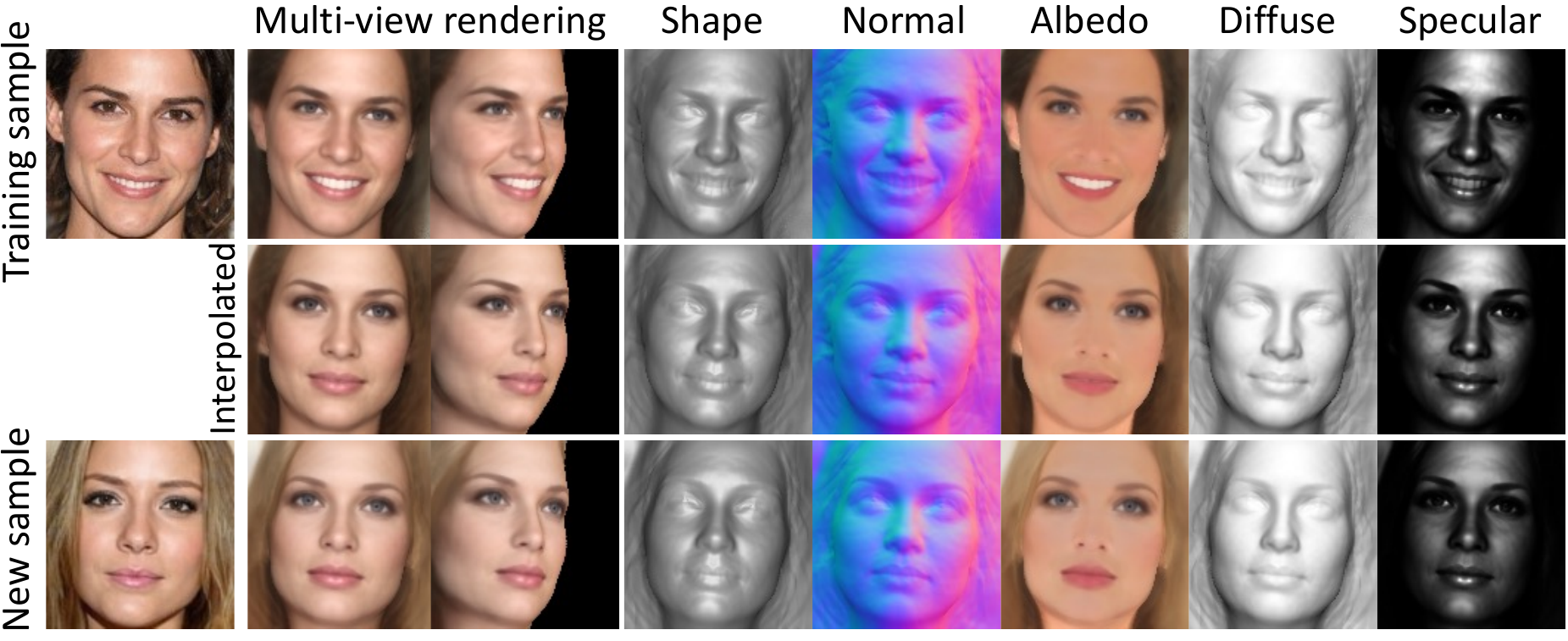}
	\vspace{-0.75cm}
	\caption{Application on lifting 2D GANs to decomposed 3D GANs.}
	\vspace{-0.35cm}
	\label{fig:generative}
\end{figure}

\noindent\textbf{Lifting 2D GANs to decomposed 3D GANs.}
Finally, we observe that GAN2X jointly trained on multiple GAN samples is capable of inheriting the generative property of the 2D GAN.
As shown in Fig.~\ref{fig:generative}, our method not only works for samples used in training, but also generalizes to new samples of the 2D GAN.
The different samples can also be naturally interpolated by interpolating the corresponding latent codes, which is an important property of GANs.
The results demonstrate the possibility of lifting a 2D GAN to a decomposed 3D GAN with our approach.

%% file: sections/conclusion.tex

\section{Conclusion}

We have presented a new approach for unsupervised inverse rendering.
To overcome the shortage of paired data, we leverage the generative modeling capability of GANs to create pseudo paired data that reflect the underlying intrinsic components.
With volume rendering and a Phong shading based implicit neural representation, our approach successfully recovers high-quality 3D shapes, albedo, and specular properties for different object categories, which opens up wide downstream applications.
Notably, GAN2X is the first method that can learn spatially-varying specular properties from merely unpaired image collections.
%
%
For future work, our approach could be further combined with recent advances in 3D GANs to further refine their geometry and decompose the material properties.

%% file: sections/appendix.tex

\clearpage


\section{Supplementary Material}

In this supplementary material, we provide the implementation details, discuss the limitations, and show more qualitative results.
We also recommend readers to refer to the video demos at the \href{https://vcai.mpi-inf.mpg.de/projects/GAN2X/}{project page}.

\subsection{Implementation Details}


\textbf{Model Architectures.}
%
Similar to the design of NeuS~\cite{wang2021neus}, our shape MLP $F_s$ has 8 hidden layers while the material MLP $F_m$ has 4 hidden layers.
The number of channels for each hidden layer is 256.
Apart from the coordinate $\vx$, $F_m$ also takes the surface normal $\vn$ and the last feature from $F_s$ as input.
Positional encoding~\cite{mildenhall2020nerf} is applied to $\vx$ with 4 frequencies for $F_s$ and 6 frequencies for $F_m$.
In volume rendering, the number of coarse and fine samples are 36 and 36 respectively.

%
The architecture for viewpoint encoder $E_v$ and lighting encoder $E_l$ is described in Tab.~\ref{tab:encoder}.
The architecture for image encoder $E_I$ and GAN encoder $E_w$ is described in Tab.~\ref{tab:encoder2}.
These architectures are based on $256^2$-resolution input images.
In our experiments, we use $256^2$ resolution for most datasets except CelebA~\cite{liu2015deep}, for which $128^2$ resolution is used.
For $128^2$-resolution input images, the second convolution layer in Tab.~\ref{tab:encoder} and the first ResBlock in Tab.~\ref{tab:encoder2} are removed while the output channel of the first convolution layer is increased from 16 to 32.
The abbreviations for the network layers are described below:

\begin{itemize}
	\item Conv($c_{in}, c_{out}, k, s, p$): convolution with $c_{in}$ input channels, $c_{out}$ output channels, kernel size $k$, stride $s$, and padding $p$.
	\item Avg\_pool($s$): average pooling with a stride of $s$.
	\item ResBlock($c_{in}, c_{out}$): residual block as defined in Tab.\ref{tab:resblock}.
\end{itemize}

\begin{table}[h!]
	\centering
	\resizebox{6.1cm}{1.8cm}{
		\begin{tabular}{lc}
			\Xhline{2\arrayrulewidth}
			Encoder & Output size \\ \hline
			Conv(3, 16, 4, 2, 1) + ReLU & 128 \\
			Conv(16, 32, 4, 2, 1) + ReLU & 64 \\
			Conv(32, 64, 4, 2, 1) + ReLU & 32 \\
			Conv(64, 128, 4, 2, 1) + ReLU & 16 \\
			Conv(128, 256, 4, 2, 1) + ReLU & 8 \\
			Conv(256, 512, 4, 2, 1) + ReLU & 4 \\
			Conv(512, 512, 4, 1, 0) + ReLU & 1 \\
			Conv(512, $c_{out}$, 1, 1, 0) + Tanh & 1 \\
			\Xhline{2\arrayrulewidth}
		\end{tabular}
	}
	\vspace{0.1cm}
	\caption{Network architecture for viewpoint net $E_v$ and lighting net $E_l$. The output channel size $c_{out}$ is 6 for $E_v$ and 4 or 5 for $E_l$ depending on whether the negative shading term is used.}\label{tab:encoder}
\end{table}

\begin{table}[h!]
	\centering
	\resizebox{6.5cm}{1.9cm}{
		\begin{tabular}{lc}
			\Xhline{2\arrayrulewidth}
			Encoder & Output size \\ \hline
			Conv(3, 16, 4, 2, 1) + ReLU & 128 \\
			ResBlock(16, 32)  & 64 \\
			ResBlock(32, 64)  & 32 \\
			ResBlock(64, 128)  & 16 \\
			ResBlock(128, 256)  & 8 \\
			ResBlock(256, 512)  & 4 \\
			Conv(512, 1024, 4, 1, 0) + ReLU & 1 \\
			Conv(1024, 512, 1, 1, 0) & 1 \\
			\Xhline{2\arrayrulewidth}
		\end{tabular}
	}
	\vspace{0.1cm}
		\caption{Network architecture of image encoder $E_I$ and GAN encoder $E_w$.}\label{tab:encoder2}
\end{table}

\begin{table}[h!]
	\centering
	\resizebox{5cm}{1.4cm}{
		\begin{tabular}{lc}
			\Xhline{2\arrayrulewidth}
			Residual path \\ \hline
			ReLU + Conv($c_{in}$, $c_{out}$, 3, 2, 1) \\
			ReLU + Conv($c_{out}$, $c_{out}$, 3, 1, 1) \\ \hline\hline
			Identity path \\ \hline
			Avg\_pool(2) \\
			Conv($c_{in}$, $c_{out}$, 1, 1, 0) \\
			\Xhline{2\arrayrulewidth}
		\end{tabular}
	}
	\vspace{0.1cm}
\caption{Network architecture for the ResBlock($c_{in}$, $c_{out}$) in Tab.\ref{tab:encoder2}. The output of Residual path and Identity path are added as the final output. }\label{tab:resblock}
\end{table}

\begin{table}[h!]
	\centering
	\resizebox{5cm}{1.9cm}{
		\begin{tabular}{cc}
			\Xhline{2\arrayrulewidth}
			Parameter & Value/Range \\ \hline
			$p$ & (4, 100) \\
			$\beta$  &  $3e^{-5}$  \\
			$\lambda_{vl}$ & 1.0 \\
			$\lambda_{d}$ & 1.0 \\
			$\lambda_{m}$ & 0.2 \\
			$lr$ for $F_s$ and $F_m$ & $5e^{-4}$ \\
			$lr$ for $F_m$ in step1 & $2e^{-3}$ \\
			$lr$ for all encoders & $2e^{-4}$ \\
			\Xhline{2\arrayrulewidth}
		\end{tabular}
	}
	\vspace{0.1cm}
	\caption{Hyper-parameters. $lr$ denotes learning rate.}\label{tab:hyperparam}
\end{table}

\begin{table}[h!]
	\centering
		\resizebox{6.3cm}{4.1cm}{
		\begin{tabular}{lc}
			\Xhline{2\arrayrulewidth}
			Joint Pre-training & Value \\ \hline
			Number of samples & (1k/200/100)  \\
			Number of re-rendered samples $m$ & (32/80/160) \\
			Number of stages  & 5  \\
			Step 1 iterations (1st stage) & 50k  \\
			Step 1 iterations (other stages) & 30k  \\
			Step 2 iterations  & 8k   \\
			Step 3 iterations  & 50k  \\
			$(\sigma_{min}, \sigma_{max})$ & (0.3, 1.0) \\
			$(\lambda_{reg1},\lambda_{reg2})$ & (0.2, 0.1)   \\ \hline\hline
			Instance-specific fine-tuning & Value  \\ \hline
			Number of re-rendered samples $m$ & (800/400/400) \\
			Number of stages  & 3  \\
			Step 1 iterations  & 6k  \\
			Step 2 iterations  & 2k   \\
			Step 3 iterations  & 15k   \\
			$(\sigma_{min}, \sigma_{max})$ & (0.8, 1.0) \\
			$(\lambda_{reg1},\lambda_{reg2})$ & (0.2, 0.1)  \\ \hline\hline
			Shading-based refinement & Value  \\ \hline
			Iterations & 1k   \\
			$(\sigma_{min}, \sigma_{max})$ & (1.0, 1.0) \\
			$(\lambda_{reg1},\lambda_{reg2})$ & (0.5, 0.2)  \\ 
			\Xhline{2\arrayrulewidth}
		\end{tabular}
	}
	\caption{Hyper-parameters for CelebA, CelebA-HQ, and AFHQ Cat datasets. (x/y/z) denote values for CelebA, CelebA-HQ, and AFHQ Cat respectively.}\label{tab:face}
\end{table}

\noindent\textbf{Hyperparameters.}
The hyperparameters used in our experiments are provided in Tab.~\ref{tab:hyperparam}, Tab.~\ref{tab:face}, and Tab.~\ref{tab:car}.
For clarity, we denote the material network optimization process in \textbf{Exploration} as ``step 1'', the GAN reconstruction process that trains $E_w$ as ``step 2'', and the \textbf{Exploitation} process as ``step 3''.
``Number of stages'' denotes how many times the exploration-and-exploitation process are repeated.
For the chromaticity-based smoothness loss, we use different $\lambda_{reg}$ values for albedo and specularity maps, which are denoted as $\lambda_{reg1}$ and $\lambda_{reg2}$ respectively.
%

\noindent\textbf{Training Process.}
For CelebA, CelebA-HQ, and AFHQ Cat datasets, we first pre-train our model on multiple samples jointly as mentioned in Sec.~3.2 of the main paper.
We then perform instance-specific training for any individual sample.
%
For LSUN car, we do not perform joint training, but first train on $128^2$ resolution and then fine-tune on $256^2$ resolution for each instance.
For all datasets, shading-based refinement is finally applied to further refine the results.
Note that for the application of lifting 2D GAN to 3D GAN, only joint pre-training is involved as there is no need for instance-specific training.
And for application on real images, only joint pre-training and shading-based refinement are involved.
This is because instance-specific fine-tuning for real images would require GAN inversion, which harms editability and thus does not perform very stable in practice.

\noindent\textbf{Losses.}
Similar to \cite{wang2021neus,yariv2020multiview}, we use an Eikonal term to regularize the SDF of $F_s$ by $\mathcal{L}_{eik} = \frac{1}{n} \sum_{i}(|\nabla F_s(\vx_i)|-1)^2$, where ${\vx_i}$ are the sampled points and the loss weight for this regularization term is 0.1.
For CelebA-HQ and LSUN car, we also include a mask loss in the same way as \cite{wang2021neus}, where the masks are obtained from off-the-shelf segmentation models.

The training process of Eq. 7 in the main paper is typically done by randomly sampling 512 pixels on the image and applying the losses with respect to these pixels.
However, the perceptual loss~\cite{zhang2018perceptual} cannot by applied to these scattered pixels and rendering the whole image is infeasible due to heavy memory consumption.
To this end, for each training iteration, we randomly choose from two pixel sampling strategies, where the first is random sampling as mentioned before and the second is to sample a $32^2$ image patch.
The first pixel sampling strategy preserves the randomness of sampled pixel positions while the second strategy allows perceptual loss to be applied to the image patch.
The perceptual loss has a loss weight of 0.1.

\begin{table}[t!]
	\centering
	\resizebox{6cm}{4cm}{
		\begin{tabular}{lc}
			\Xhline{2\arrayrulewidth}
			Pre-train on 128 resolution & Value \\ \hline
			Number of re-rendered samples $m$ & 800 \\
			Number of stages  & 4  \\
			Step 1 iterations (1st stage) & 15k  \\
			Step 1 iterations (other stages) & 10k  \\
			Step 2 iterations  & 4k   \\
			Step 3 iterations  & 25k  \\
			$(\sigma_{min}, \sigma_{max})$ & (0.3, 1.0) \\
			$(\lambda_{reg1},\lambda_{reg2})$ & (0.2, 0.1)   \\ \hline\hline
			Fine-tune on 256 resolution & Value  \\ \hline
			Number of re-rendered samples $m$ & 400 \\
			Number of stages  & 3  \\
			Step 1 iterations  & 6k  \\
			Step 2 iterations  & 2k   \\
			Step 3 iterations  & 15k   \\
			$(\sigma_{min}, \sigma_{max})$ & (0.8, 1.0) \\
			$(\lambda_{reg1},\lambda_{reg2})$ & (0.2, 0.1)  \\ \hline\hline
			Shading-based refinement & Value  \\ \hline
			Iterations & 1k   \\
			$(\sigma_{min}, \sigma_{max})$ & (1.0, 1.0) \\
			$(\lambda_{reg1},\lambda_{reg2})$ & (0.5, 0.2)  \\ 
			\Xhline{2\arrayrulewidth}
		\end{tabular}
	}
	\caption{Hyper-parameters for LSUN car dataset. }\label{tab:car}
\end{table}

%

\noindent\textbf{Other Training Details.}
Our implementation is based on PyTorch~\cite{paszke2017automatic}. We use Adam optimizer~\cite{kingma2014adam} in all experiments.
The joint pre-training is run on 2 RTX 8000 GPUs, while all other instance-specific trainings are run on 1 RTX 8000 GPU.

For CelebA, CelebA-HQ, and AFHQ Cat datasets, we adopt a symmetry assumption on object shape and material at step3 of the first stage in joint pre-training.
This is done by randomly flipping the shape and material during training, which is similar to \cite{wu2020unsupervised}.
This symmetry assumption helps to infer a canonical face pose.


Note that the exploration process of our method involves randomly sampling multiple viewpoints and lighting conditions.
Here we follow \cite{pan2020gan2shape}, where the viewpoints are sampled from a prior multi-variate normal distribution and the lighting conditions are sampled from a prior uniform distribution.



\begin{figure*}[t!]
\centering
\includegraphics[width=16.5cm]{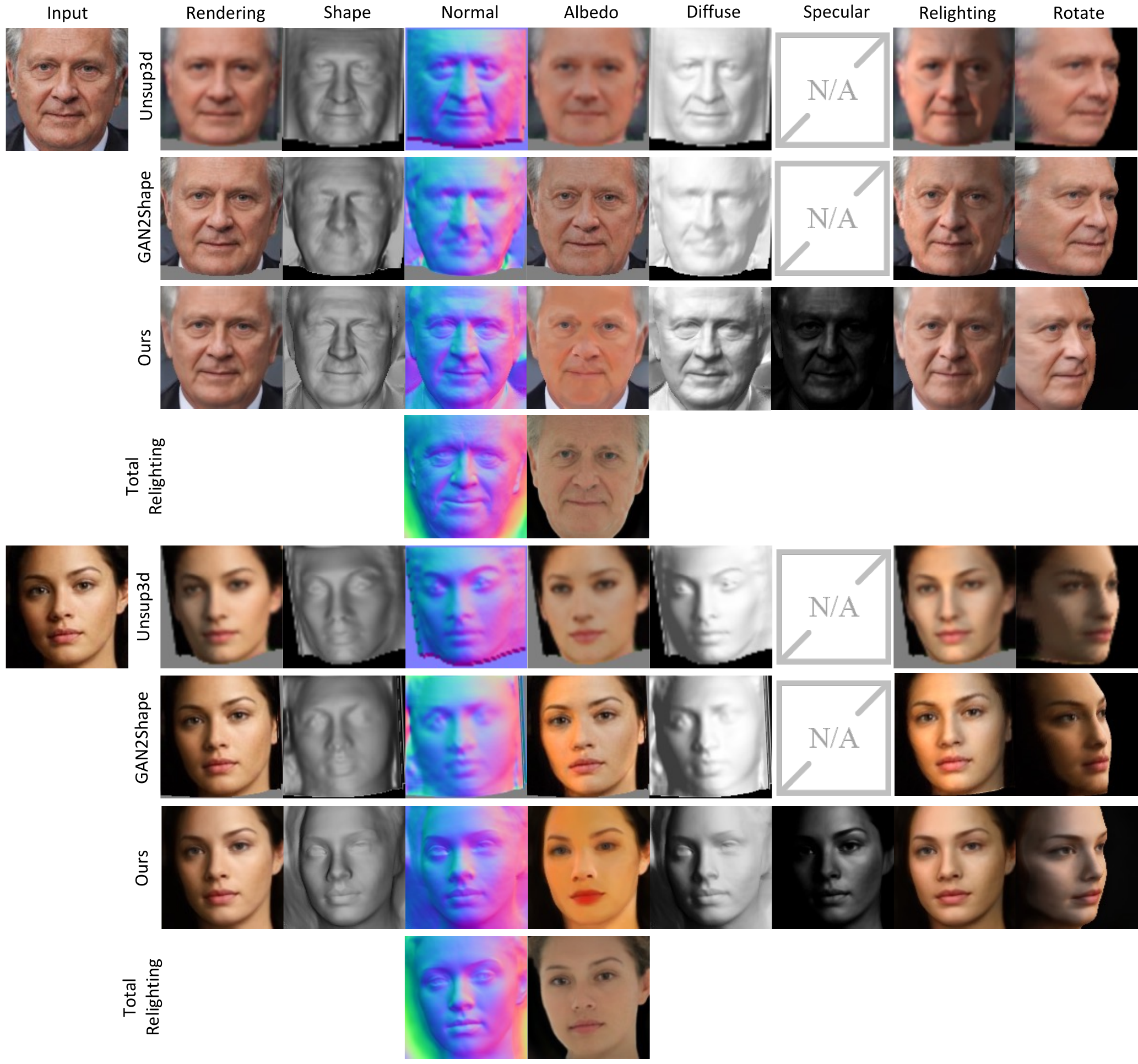}
\caption{More qualitative comparison. This is an extension of Fig. 4 in the main paper.}
\label{fig:qualitative_comp_sup}
\end{figure*}

\vspace{5pt}
\subsection{Limitations}
While our approach shows promising inverse rendering results, it also has some limitations.
First, we assume simplified lighting to reduce ambiguity and to make the problem tractable, which may not be sufficient to well approximate more complex lighting.
Besides, as the exploration step in our approach relies on a convex shape prior, it mainly works for roughly convex objects and is hard to be applied to more complex objects (\eg, bicycles).
This might be alleviated via the recent advancement in 3D-aware GANs that allow explicit camera pose control.

\vspace{5pt}
\subsection{Qualitative Results}

\begin{figure*}[h!]
\centering
\includegraphics[width=14.5cm]{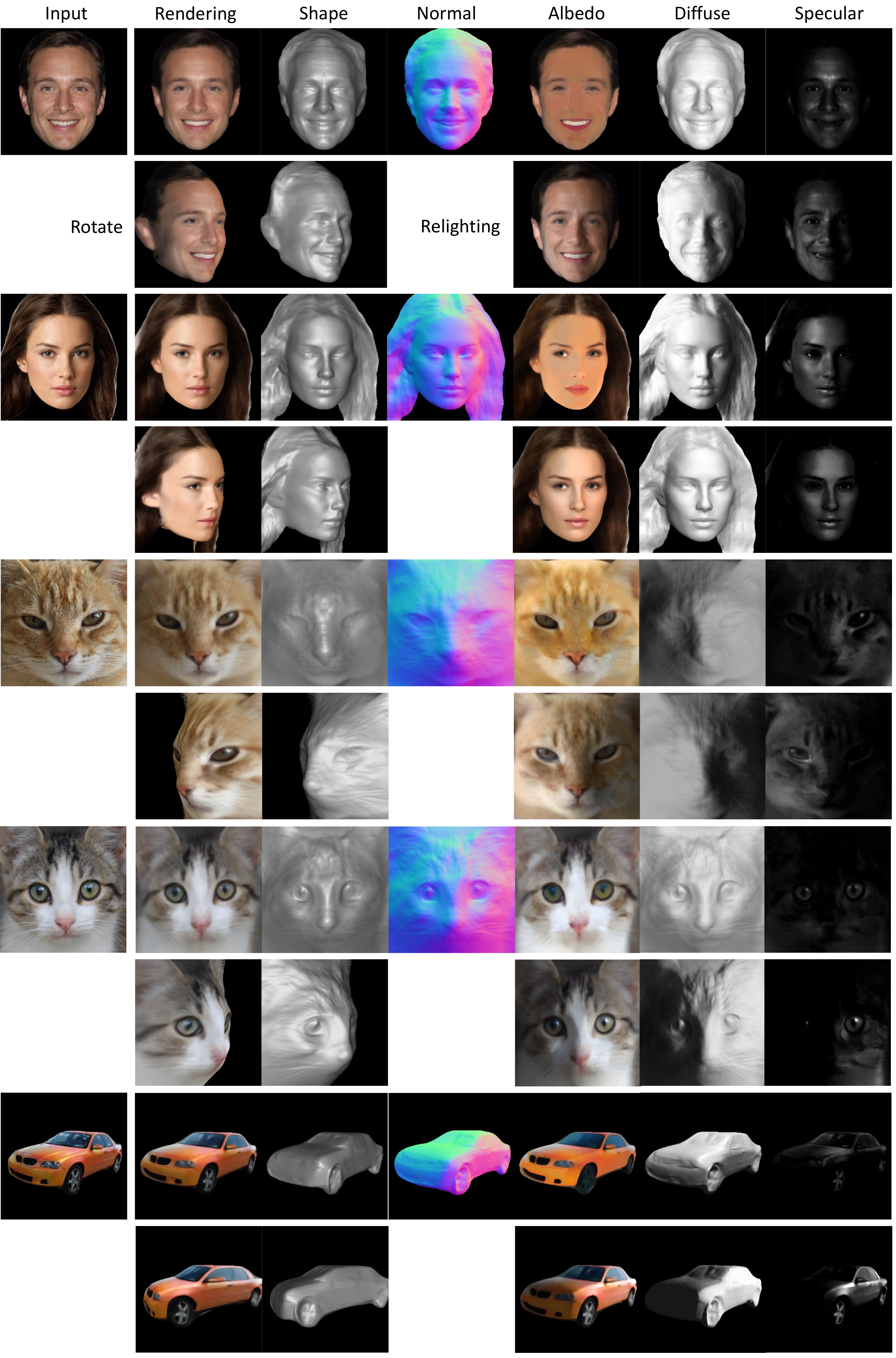}
\vspace{-0.1cm}
\caption{More qualitative results. This is an extension of Fig. 5 in the main paper.}
\vspace{-0.1cm}
\label{fig:qualitative_sup}
\end{figure*}

In this section, we provide more qualitative results of our method.
Fig.~\ref{fig:qualitative_comp_sup} provides more qualitative comparison with Unsup3d~\cite{wu2020unsupervised} and GAN2Shape~\cite{pan2020gan2shape}.
We also show the albedo and normal of the supervised method Total Relighting~\cite{pandey2021total} as a reference.
More qualitative results are shown in Fig.~\ref{fig:qualitative_sup}.

\begin{figure*}[h!]
\centering
\includegraphics[width=12cm]{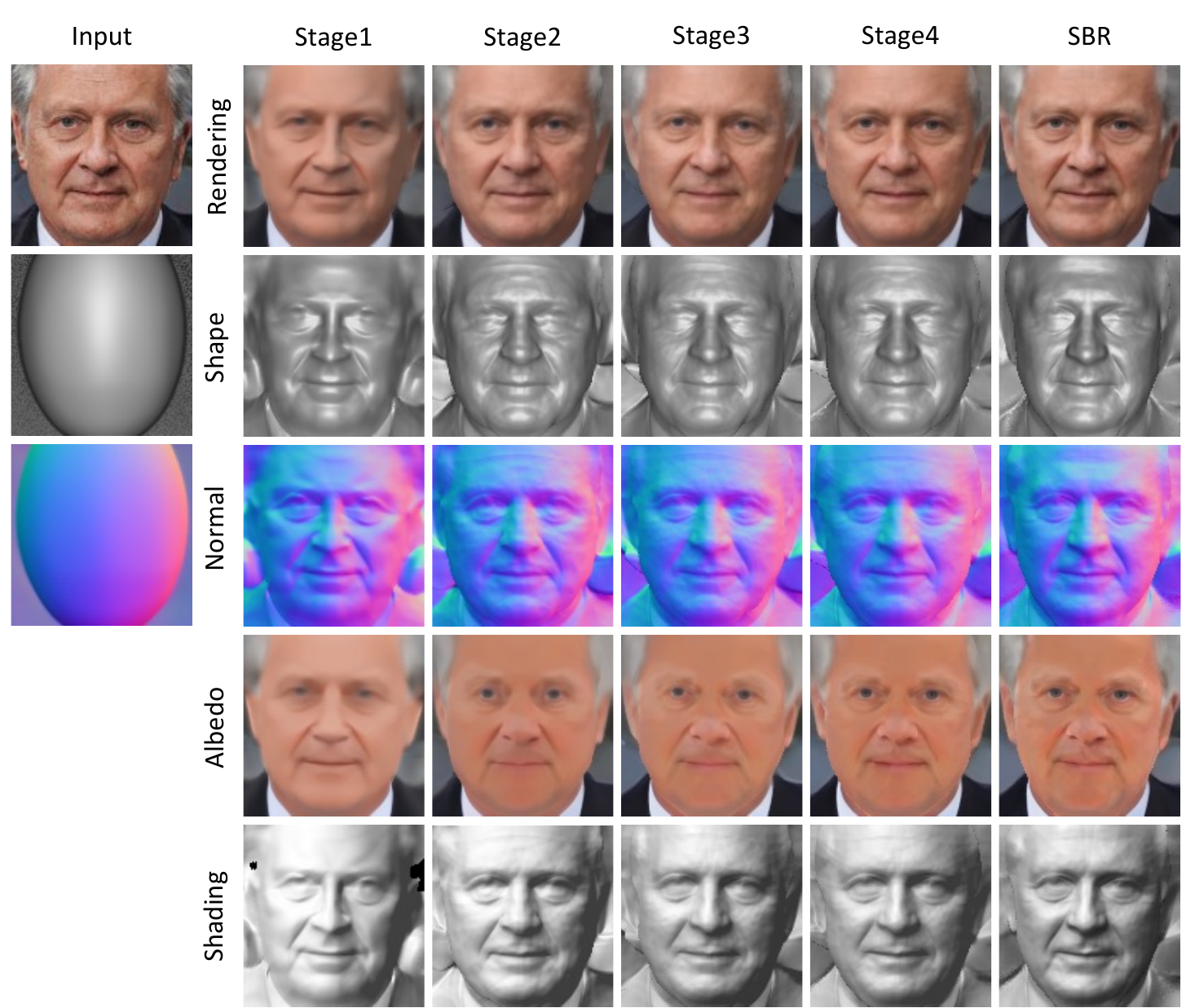}
\caption{Effects of progressive training and shading-based refinement (SBR).}
\label{fig:progress}
\end{figure*}

\begin{figure*}[h!]
\centering
\includegraphics[width=12cm]{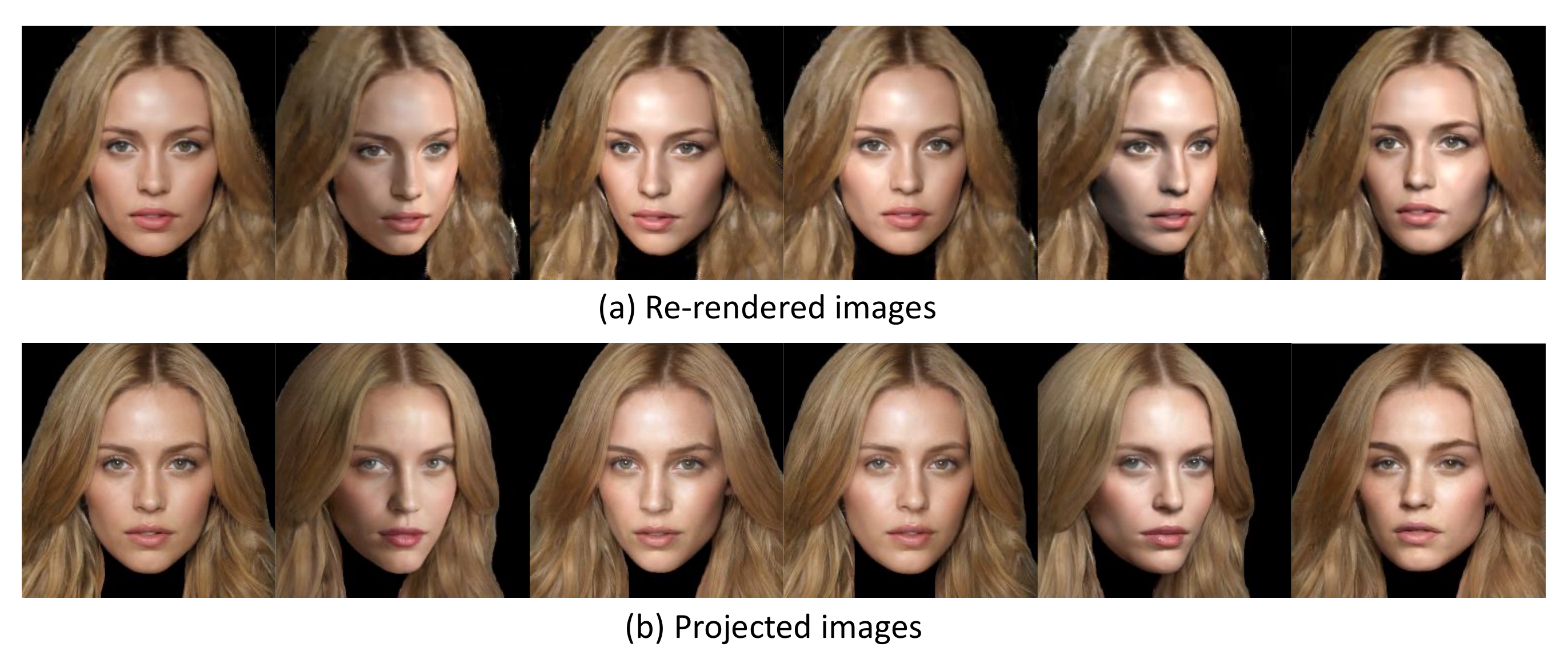}
\caption{Examples of (a) re-rendered images and (b) their corresponding projected images.}
\label{fig:samples}
\end{figure*}

Note that our method repeats the exploration-and-exploitation process for several stages.
We show the effects of this progressive training in Fig.~\ref{fig:progress}.
It can be seen that the results get more accurate with more training stages.
The shading-based refinement further refines the results to be more precise.
In Fig.~\ref{fig:samples}, we provide some examples of re-rendered images and projected images during training. It can be observed that the projected images have similar viewpoints and lighting conditions as the re-rendered images, but are more natural and thus provide useful information to refine the object intrinsics.

